# Spectral-Spatial Classification of Hyperspectral Image Using Autoencoders

Zhouhan Lin, Yushi Chen, Xing Zhao
Dept. of Information Engineering
Harbin Institute of Technology
Harbin, China
lin.zhouhan@gmail.com, chenyushi@hit.edu.cn,
xintongzhaoxing@126.com

Gang Wang
Sch. of Electrics & Electronics Engineering
Nanyang Technological University
Singapore, Singapore
wanggang@ntu.edu.sg

*Abstract*—**Hyperspectral image (HSI) classification is a hot topic in the remote sensing community. This paper proposes a new framework of spectral-spatial feature extraction for HSI classification, in which for the first time the concept of deep learning is introduced. Specifically, the model of autoencoder is exploited in our framework to extract various kinds of features. First we verify the eligibility of autoencoder by following classical spectral information based classification and use autoencoders with different depth to classify hyperspectral image. Further in the proposed framework, we combine PCA on spectral dimension and autoencoder on the other two spatial dimensions to extract spectral-spatial information for classification. The experimental results show that this framework achieves the highest classification accuracy among all methods, and outperforms classical classifiers such as SVM and PCA-based SVM.**

## I. INTRODUCTION

By combining imaging and spectroscopy technology, hyperspectral remote sensing can get spatially and spectrally continuous data simultaneously. Hyperspectral imagery is becoming a valuable tool for monitoring the Earth's surface [1]. If successfully exploited, the hyperspectral image can yield higher classification accuracies and more detailed class taxonomies [2].

Traditional HSI classification methods uses spectral information only, and the classification algorithms typically include parallelepiped classification, *k*-nearest-neighbors, maximum likelihood, minimum distance and logistic regression [3]. Since HSI classification deals with a problem with high-dimensional feature space and low number of labeled data, the majority of these above algorithms suffer a lot from "curse of dimensionality."[4] To tackle this problem, scholars of the remote sensing community try to introduce variety of feature extraction methods like PCA, ICA, sequential forward floating search and wavelet analysis [5]. However, these methods do not bring significant improvement for classification, in some cases the resulting accuracies even fall below direct-classification approaches.

However, large marginal machines like SVM deals well on the above over fitting problem. In most cases, SVM based methods can obtain better classification accuracy than other widely used pattern recognition techniques on HSI data [3]. So after SVM is introduced in this field [6], feature extraction is seldom used, and applying SVM based method on the original data becomes the state-of-the-art method in HSI classification. However, the ways all these generally used methods are applied have one thing in common: They deal with spectral information of each pixel only. But as recently the resolution of imaging spectrometer burgeons, the spatial information seems growing more important for further improving HSI classification accuracy. Recently there are some methods proposed for incorporating spatial information, like mathematical morphology [7].

*This work is supported by "the Fundamental Research Funds for the Central Universities" (Grant No. HIT. NSRIF.2013028) and National Natural Science Foundation of China( No. 61301206 )

*Codes for the Stacked Autoencoder models mentioned in this paper are available at
https://github.com/hantek/deeplearn_hsi

On the other thread, recent advantages in training multilayer neural networks have refurbished best records in a wide variety of machine learning problems including classification or regression tasks that involve processing image [8], language [9] and speech [10]. Typical deep neural network architectures include Deep Belief Networks [11], Deep Boltzmann Machines [12], Stacked Autoencoders [13] and Stacked Denoising Autoencoders [14]. The layer-wise training framework has a bunch of alternatives like Restricted Boltzmann Machines [15], Pooling Units [16], Convolutional Neural Networks [17] and Autoencoders [13].

In this paper, we introduce deep learning based feature extraction for HSI classification for the first time. Our work focuses applying one of the aforementioned models – autoencoders – to learn representations (i.e. features) of hyperspectral image data in an unsupervised manner. Since hidden layer activities are exploited as features of the initial data, and they are called "representations" in the neural network vocabulary, we will use the word "representation" instead of "feature" in the rest of this paper. Our methods exploit single-layer autoencoder and multi-layer stacked autoencoder to extract shallow and deep representations of hyperspectral image data respectively. Further we propose a new way of extracting spectral-spatial representation for classification. These representations are then exploited to deal with HSI classification problems. The effectiveness of these methods is verified by classification accuracy.

The rest of this paper is organized in four sections. Section II is a description for the hyperspectral image classification task we are dealing with, and a terse introduction on the autoencoder models used in this paper. Section III details on how we extract various representations with the novel architecture. Subsequent classification is conducted by logistic regression and SVM. Section IV deals with the experimental results. Finally, Section V summarizes the observations and point out some probable future works to complete this paper.

## II. BACKGROUND

### A. *Hyperspectral Image Classification: Description*

Classifying hyperspectral image is a little bit different with ordinary image classification. So we need to elaborate here the task we are facing.

A typical scene of hyperspectral image covers several square kilometers of lands and has hundreds of spectral channels instead of only 3 RGB channels, as shown in Fig 1. As a result, the task of classifying hyperspectral image has unique characters. The most unique one is that, instead of giving one label to a whole image, hyperspectral image classification task is pixel-based: We arrange a label for each pixel in the scene according to the spectral information provided by its hundreds of spectral channels.

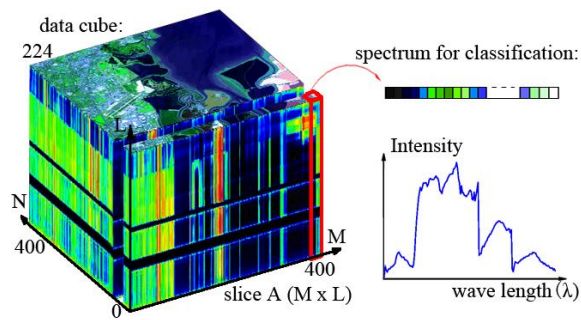

Figure 1. A typical scene of hyperspectral image. Each pixel consists of a whole spectrum.

## B. Autoencoders (AE) and Stacked autoencoders (SAE)

An autoencoder (AE) has one visible layer of $d$ inputs and one hidden layer of $h$ units with an activation function $f$. During training, it first maps the input $x \epsilon R^d$ to the hidden layer and get the latent representation $y \epsilon R^h$; and then $y$ is mapped to an output layer that has the same size with input layer, which is called "reconstruction." The reconstruction is denoted as $z \epsilon R^d$ (Fig 2).

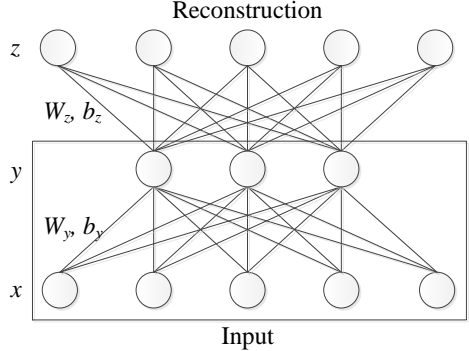

Figure 2. A single layer autoencoder for HSI classification. The model learns a hidden representation "$y$" from input "$x$" by reconstructing it on "$z$". Corresponding parameters are denoted in the network.

Mathematically, these procedures can be shown as:

$$y = f(W_y x + b_y) \quad (1)$$

$$z = f(W_z y + b_z) \quad (2)$$

where $W_y, W_z$ denotes input-to-hidden and hidden-to-output weights respectively, $b_y, b_z$ denotes the bias of hidden and output units, and $f(\cdot)$ denotes the activation function, which apply element-wise to its arguments. The goal of training is to minimize the "error" between input and reconstructed input, i.e.,

$$\underset{W, b_y, b_z}{\operatorname{argmin}}[c(x,z)] \quad (3)$$

where $z$ is dependent on parameters $W, b_y, b_z$ while $x$ is given. $c(x,z)$ stands for the "error," which can be defined in a variety of ways.

We introduce stacked autoencoder (SAE) to help us compute deep representations of spectral data. The SAE has multiple layers of autoencoders and yields a deep representation of input data at the output of the last layer. It is trained in a layer-wise manner. That is, after we finish training a former layer of parameters, subsequent layer is trained according to the output of its previous layer. Stacking these input-to-hidden layers sequentially constructs a stacked autoencoder.

## III. LEARNING REPRESENTATIONS FOR HSI CLASSIFICATION

Our methods involve feature extraction for HSI classification. We first compute representations via autoencoder and deem it as the feature of data, then construct a classifier on the neural network to finish the classification phase. This section focuses on how the varieties of features are extracted and incorporated in classification. For classical HSI classification which exploits spectral information only, we first propose two kinds of classification schemes exploiting shallow and deep representations of HSI spectra respectively. In the last part of this section, we propose a novel classification framework which exploits both spectral and spatial information.

### A. Classifying with shallow spectral representation

This scheme consists of two steps. First an autoencoder is used to extract single-layer spectral representation and then a SVM is constructed on top of the hidden layer of autoencoder.

In the autoencoder layer, the activation function in our method is set to be sigmoidal, i.e. $f(x) = \frac{1}{1+e^{-x}}$. Its first-order and second-order derivative has special forms that brings convenience for computing: $f'(x) = f(x)[1 - f(x)]$, $f''(x) = f(x)[1 - f(x)][1 - 2f(x)]$. And to reduce the number of parameters, we use tied weights while we are training the autoencoder, i.e. let $W_y = W_z = W$. So now there's 3 groups of parameters remaining to learn: $W, b_y$ and $b_z$.

We choose the cross entropy as the cost function here, which is in accordance with the sigmoid output. In our implementation, the cost is computed on a minibatch of inputs since we adopt minibatch update strategy for the large dataset.

$$c = -\frac{1}{m}\sum_{i=1}^{m}\sum_{k=1}^{d}[x_{ik} \log(z_{ik}) + (1 - x_{ik}) \log(1 - z_{ik})] \tag{5}$$

Here $d$ denotes the input vector size, and $m$ denotes minibatch size. $x_{ik}(z_{ik})$ denotes $k$-th element of the $i$-th input (reconstruction) in the minibatch. The inner summation is over the input dimension, while the outer over a whole minibatch.

Our hope turns to optimize Eq. 5 using minibatch stochastic gradient descent. We'll now derive the partial differentials of cost with respect to parameters $W, b_y$ and $b_z$. First we rewrite reconstruction in a scalar form:

$$net_{ip}^{y} = \sum_{q=1}^{d} x_{iq} W_{qp} + b_{yp} \tag{6}$$

$$net_{ik}^{z} = \sum_{p=1}^{h} W_{kp} f(net_{ip}^{y}) + b_{zk} \tag{7}$$

$$z_{ik} = f(net_{ik}^{z}) = f\left(\sum_{p=1}^{h} W_{kp} f\left(\sum_{q=1}^{d} x_{iq} W_{qp} + b_{yp}\right) + b_{zk}\right) \tag{8}$$

where $net_{ip}^{y}$, $net_{ik}^{z}$ denotes the net input of the $p$-th hidden and output unit, given the $i$-th sample in the minibatch.

Putting them all together (Eqs. 5~8), and using the chain rule, we have partial differentials of cost (Eq. 5) over parameters $W, b_y$ and $b_z$.

$$\begin{cases} \frac{\partial c}{\partial W_{rs}} = -\frac{1}{m}\sum_{i=1}^{m}\left\{\sum_{k=1}^{d}\left[\frac{x_{ik} - z_{ik}}{z_{ik}(1 - z_{ik})}f'(net_{ik}^{z})W_{ks}f'(net_{is}^{y})x_{ir}\right] + f'(net_{ik}^{z})f(net_{is}^{y})\right\} \\ \frac{\partial c}{\partial b_{yr}} = -\frac{1}{m}\sum_{i=1}^{m}\sum_{k=1}^{d}\frac{x_{ik} - z_{ik}}{z_{ik}(1 - z_{ik})}f'(net_{ik}^{z})W_{kr}f'(net_{ir}^{y}) \\ \frac{\partial c}{\partial b_{zr}} = -\frac{1}{m}\sum_{i=1}^{m}\sum_{k=1}^{d}\frac{x_{ik} - z_{ik}}{z_{ik}(1 - z_{ik})}f'(net_{ik}^{z}) \end{cases} \tag{9}$$

Eq. 9 gives explicitly the gradients we need in conducting stochastic gradient descent, by which we learn the 3 groups of parameters.

After training the network, we remove the reconstruction layer and deem the hidden activity to be the learned representation. The second step is to put an SVM with linear kernel on top of its hidden layer.

What we should notice is that the two models are trained separately, i.e., the SVM is trained after all parameters of the autoencoder are determined, and training SVM does not change anything of the autoencoder. So this classification scheme should be called Autoencoder-based SVM, and we will call it AE-SVM for short in the rest part of the paper.

*B. Classifying with deep spectral representation*

There exist some motivations to extract more robust deep spectral representations. First, because of the complex situation of lightening in the large scene, objects in a same class show different spectral characters in different location. For example, a lawn exposed in direct sunshine shows different spectral characters from a same lawn but eclipsed from the sunshine by a high building. Also, scattering from other peripheral ground objects tilts the spectra of the lawn and change its characters too. Other factors involve rotations of the sensor, different atmospheric scattering condition, and so on. According to these factors, the probability distribution of a certain class is hard to be one-hot and has variations over multiple directions in the feature space. These complex variations of spectra make it hopeless to analyze pixel by pixel how they are affected by their tangent pixels in the complicated real situation, thus they demand more robust and invariant features.

It is believed that deep architectures can potentially lead to progressively more abstract features at higher layers of representation, and more abstract features are generally invariant to most local changes of the input [17]. So, to get more generally invariant representations and tackle these problems, stacked autoencoder and its corresponding deep spectral representations are suitable for this problem.

We introduce stacked autoencoder to help us get deep representations of spectral data, and add a logistic regression classifier on top of the stacked autoencoder to reach the high classification accuracy. Fig 3 shows a typical instance of the deep architecture used in our paper. The first layer autoencoder maps inputs in layer 0 to a first layer representation in layer 1. It is trained in the same manner to aforementioned single layer autoencoder.

For logistic regression, we use soft-max as its output layer activation, and output layer size is the same to total number of classes. Since it is implemented as a single-layer neural network, it is merged with the former layers of network. Fitting the classifier is conducted over the whole architecture, but with very slight learning rates on former layer autoencoders.

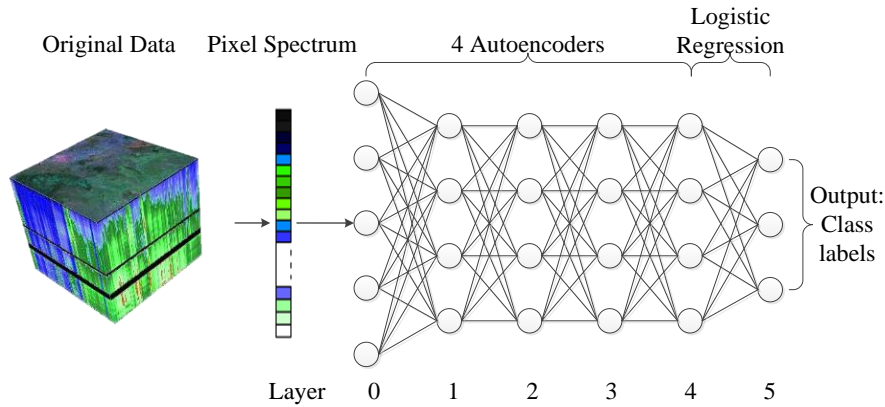

Figure 3. Classifying with deep spectral representation. This instance of classification scheme shown here has 6 layers: one input layer, 4 hidden layers of autoencoders and an output layer of logistic regression.

*C. Classifying with spectral-spatial representation*

Unlike other HSI spectral-spatial information extraction methods which only use the 4 or 8 tangent neighbors and simple filtering, our deep framework takes all the pixels in a flat neighbor region into consideration, and let the autoencoders learn the representation by itself. The overall flowchart of our proposed method is detailed in Fig 4.

First a 7x7 neighbor region of a certain pixel is extracted from the original image. Due to the hundreds of channels along the spectral dimension, data on this initial layer always have tens of thousands of dimensions. Such a large neighbor region will result in too large an input dimension for the classifier, and contains too large amount of redundancy.

In the second layer, PCA is introduced to condense the spectral information, thus reducing data dimension to an acceptable scale, but reserve spatial information at the meantime. Since we mainly care about incorporating spatial information in this method, we use PCA along the spectral dimension and only retain the first 3 principle components. The PCA transformation matrix is fitted on the whole image, both for tagged and untagged pixels. This step does cast away part of spectral information, but since PCA is conducted within pixels, the spatial information remains intact.

After these processing, we "flatten" the data in the third layer, i.e., stench it onto a 1-D vector, and feed it into a stacked autoencoder.

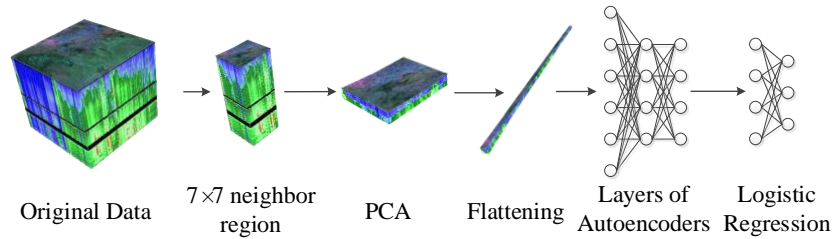

Original Data → 7×7 neighbor region → PCA → Flattening → Layers of Autoencoders → Logistic Regression

Figure 4. Spectral-spatial representation extraction scheme. The first step of procesing is PCA compressing over spectral dimension, then after flatening the data, autoencoders are introduced to extract layer-wise deep representations.

The subsequent layers include layer-wise training autoencoders and fine-tune the whole model with a final logistic regression layer. These steps are similar to the former subsection which deals with deep spectral representation and thus we would not repeat describing them here.

## IV. Experimental Results

In our study, two hyperspectral images with different environmental settings are applied to validate the proposed autoencoder-based classification method. The first is a mixed vegetation site over Kennedy Space Center (KSC), Florida, which has 13 land cover categories, 224 spectral channels. Due to water absorption and existence of low signal-noise ratio channels, 176 of them are used in our experiments. It has $512 \times 614$ pixels, in which 5211 pixels are labeled. The second is the urban site over the city of Pavia, Italy, which has 9 land cover classes, 103 spectral channels. Its spatial size is $512 \times 614$, with 42469 pixels labeled.

Our experiments are three-folded, using the aforementioned three methods respectively. All data are regularized on to an interval of [0, 1] before feeding into classifiers. We conduct our experiments on an Ubuntu 12.10 system, on an Intel i5-3230M processor, which has 2 cores of 2.6GHZ. The codes are implemented using Theano [18], a Python library for symbolic computation. The SVM and PCA are implementations from SciKit-Learn, a Python machine learning package.

### A. Single-layer spectral representation

This part compares our proposed AE-SVM classification scheme with both ordinary SVM and SVM with PCA plugged in as feature extraction step (PCA-SVM).

#### 1) Reconstructed Spectrum

Since single layer autoencoder is an important building block of other methods proposed in this paper, we first inspect on how well the input is reconstructed through epochs before we give resulting classification accuracies by examine one sample on different training epochs. The examined autoencoder has 100 hidden units and trained on KSC data. It is shown that the autoencoder restitutes a really perfect reconstruction from several hundreds of iterating epochs (Fig. 5).

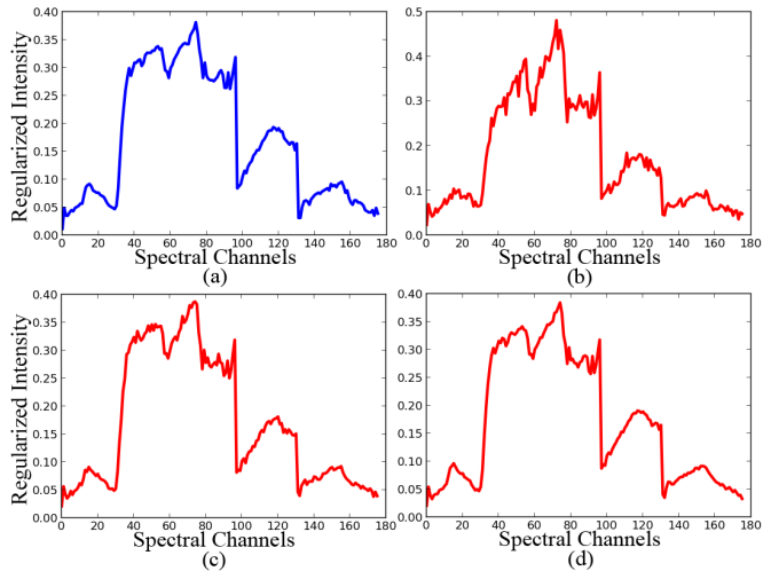

Figure 5. Reconstructions of a same input in different iteration epochs. (a) Input spectrum. (b) ~ (d) Reconstruction of (a) in epoch 1, 100 and 1000 respectively.

*2) Classification accuracy*

Fig. 6 shows the classification performance of 8 autoencoders with different sizes of hidden layers ranging from 20 to 140 on KSC data set and 60 to 180 on Pavia respectively. Although the hidden layer sizes range widely, the performance of the corresponding autoencoder-based SVM always outperforms direct applying SVM after an abundant number of iterations. However, if we accept PCA as a feature extraction step before we conduct SVM (PCA-SVM), then no matter how many principle components are incorporated, the performances of PCA-SVM are never better than direct SVM. Our experiments show that autoencoders help a lot in improving SVM classification performance. And in addition, unlike multilayer perceptron, the result is not sensitive to the number of hidden units.

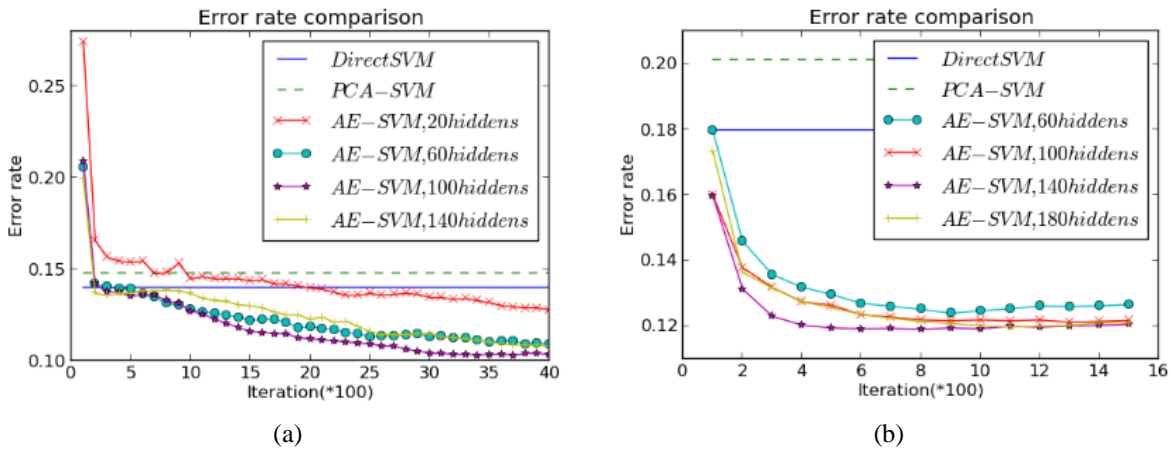

Figure 6. AE-SVM performance w.r.t. training epochs on (a) KSC data and (b) Pavia data. With appropriately large iteration epochs, the autoencoder will convergr to useful representations for SVM. And as shown in these figures, AE-SVMs do not show considerable sensitivity to the hidden layer size.

*B. Classifying with deep spectral representation*

We tried several stacked autoencoders with different depths and compare the results with other traditional methods. For KSC data, it has 176 spectral channels, and each hidden layer size is set to be 100, plus a logistic regression layer on top of the stacked autoencoder. So the neural networks are constructed as 176-100-…-100-13. For Pavia dataset who has 103 spectral channels and 9 classes, the performance reaches its best when using 140 as its hidden layers size. The neural networks are like 103-140-…-140-9.

"Hidden layer numbers" in Table I correspond to the number of 100 or 140-sized layers in the deep neural network. The experiments show that depth does help descending classification error rate. For comparison, we conduct 2 traditional methods, SVM and SVM with PCA plugged in as feature extraction, on the same data. It has shown a significant gain of accuracies on both of the images. (Table I)

TABLE I. CLASSIFICATION WITH DEEP SPECTRAL REPRESENTATION

| Deep Classifiers | Overall test set error rate (%) | |
|---|---|---|
| | KSC | Pavia |
| SAE-LR, 1 hidden layer | 9.886 | 20.115 |
| SAE-LR, 2 hidden layers | 7.829 | 17.678 |
| SAE-LR, 3 hidden layers | 7.314 | 16.138 |
| SAE-LR, 4 hidden layers | **6.229** | **14.851** |
| Control group 1: PCA-SVM | 15.102 | 20.114 |
| Control group 2: SVM | 13.971 | 17.975 |

*C. Classifying with spectral-spatial representation*

If we directly apply SVM on the spectral-spatial information, the error rate will be unacceptably high, as shown in Table II. Our methods confirm that this kind of representation also provides abundant information for classification. We use only the former 3 principle components of a 7x7 neighbor region to reach an error rate as low as 4.000% on KSC data and 14.355% on Pavia data.

Classifiers in the control groups are the same as the former subsection, and are conducted on the PCA-compressed data. However, these traditional methods fail in yielding good enough accuracy; but our proposed method succeeds in finding correct features in the dataset and yields the highest accuracies.

TABLE II. CLASSIFICATION WITH DEEP SPECTRAL-SPATIAL REPRESENTATION

| Deep Classifiers | Overall test set error rate (%) | |
|---|---|---|
| | KSC | Pavia |
| 25872(15141)-147-310-13(9) | 11.524 | 18.473 |
| 25872(15141)-147-310-100-13(9) | 6.476 | 15.704 |
| 25872(15141)-147-310-100[*2]-13(9) | 4.762 | 14.402 |
| 25872(15141)-147-310-100[*3]-13(9) | **4.000** | **14.355** |
| Control group 1: PCA-SVM | 32.191 | 21.645 |
| Control group 2: SVM | 26.286 | 20.213 |

Notes: 100[*3] denotes for 3 layers with each sized 100.
Numbers in parentheses denotes the neural network size on Pavia dataset in the corresponding layer.

V. CONCLUSION AND DISCUSSION

In this paper, we propose a hyperspectral image classification framework using both spectral and spatial information extracted by stacked autoencoders.

Our experiments first confirm that autoencoder-extracted representations help lowering error rate of SVM, a classical classifier previously considered as state-of-the-art in this field. It is shown that autoencoders are not sensitive to hidden unit number. What's more, the impact that the depth of representation has on classifying hyperspectral images is also inspected, experiments suggest that deeper representations always lead to better classification accuracies. For spectral-spatial information based classification, the proposed deep framework performs well and succeeded in classifying hyperspectral images with highest accuracies. Our future work involves incorporating other kinds of deep learning models into this framework to further improve the classification accuracy.